\newcommand{\fref}[1]{Fig.~\ref{#1}}

\newcommand{\sref}[1]{Section~\ref{#1}}

\documentclass[letter,journal]{IEEEtran}
\usepackage{amsmath,amsfonts,bbm}
\usepackage{algorithmic}
\usepackage{array}
\usepackage{stfloats}
\usepackage{url}
\usepackage{verbatim}
\usepackage{graphicx}
\usepackage[margin=0.485in]{geometry}
\usepackage{xcolor}
\usepackage{cancel}

\def\BibTeX{{\rm B\kern-.05em{\sc i\kern-.025em b}\kern-.08em
    T\kern-.1667em\lower.7ex\hbox{E}\kern-.125emX}}

\begin{document}

\title{\huge Digital Twin Graph: Automated Domain-Agnostic Construction, Fusion, and Simulation of IoT-Enabled World}
\author{Jiadi Du, Tie Luo, {\it Senior Member, IEEE}
\thanks{J. Du and T. Luo are with the Department of Computer Science, Missouri University of Science and Technology, Rolla, USA. E-mail: \{j.du, tluo\}@mst.edu. %Tie Luo is the corresponding author.
}\vspace{-8mm}}

% \markboth{Journal of \LaTeX\ Class Files,~Vol.~18, No.~9, September~2020}%
% {How to Use the IEEEtran \LaTeX \ Templates}

\maketitle

\begin{abstract}
% This document describes the most common article elements and how to use the IEEEtran class with \LaTeX \ to produce files that are suitable for submission to the Institute of Electrical and Electronics Engineers (IEEE).  IEEEtran can produce conference, journal and technical note (correspondence) papers with a suitable choice of class options.
With the advances of IoT developments, copious sensor data are communicated through wireless networks and create the opportunity of building Digital Twins to mirror and simulate the complex physical world. Digital Twin has long been believed to rely heavily on {\em domain knowledge}, but we argue that this leads to a high barrier of entry and slow development due to the scarcity and cost of human experts. In this paper, we propose Digital Twin Graph (DTG), a general data structure associated with a processing framework that constructs digital twins in a fully automated and domain-agnostic manner. This work represents the first effort that takes a completely data-driven and (unconventional) graph learning approach to addresses key digital twin challenges. %Data and system heterogeneity are also taken into full account.
%Our DTG consists of a novel Graph2Graph (G2G) transformation model, .... In addition, it can fuse (but not rely on) domain knowledge when available, using a knowledge distillation model called GAEN, and perform system-wide simulation using a graph neural network (GNN) based strategy. 

% \noindent{\blue{Making the contribution looks easy to understand but conspicuous}}
\end{abstract}

\begin{IEEEkeywords} Digital Twin, Graph neural networks, Internet of Things, Generative adversarial ensemble network. \end{IEEEkeywords}

\vspace{-3mm}
\section{Introduction}

\IEEEPARstart{O}{ver} decades, high-fidelity simulation of physical systems has been constantly sought after. Recently, with the advances of Internet of Things (IoT) for sensing, wireless communication, and the surge of computing power, {\em Digital Twin} has emerged and been envisaged as a very promising and most advanced tool for monitoring and simulating of the physical world. While its origin could be dated back to as early as the 1990's, % \cite{gelernter1993mirror}
it was first publicly introduced as a conceptual model underlying {\em product lifecycle management} at a conference on manufacturing %Society of Manufacturing Engineers 
in 2002 and republished in 2019 \cite{grieves2019virtually}. The NASA 2010 Roadmap Report described digital twin as three distinct parts: the physical world, digital representation, and communication between the two \cite{piascik2010draft}. In 2018, the U.S. DoD Digital Engineering Strategy Initiative formulated digital twin as ``An integrated multiphysics, multiscale, probabilistic simulation of an as-built system, enabled by Digital Thread, that uses the best available models, sensor information, and input data to mirror and predict activities/performance over the life of its corresponding physical twin''.
% ~\cite{dauGlossary}
Thus, digital twin bridges wireless sensor communication networks and physical system modeling and simulation.

The recent digital twin research has primarily focused on the connection models between physics and its corresponding virtual representation in different domains. Due to the complexity of the physical world, most of the studies either (1) limit the scale of digital twin to a single entity (e.g.,  ‘vehicle’, ‘component’, ‘product’ depending on specific domains) or (2) treat the whole complex virtual twin as a black box and only train a predictive model of target values using multivariate sequential data (e.g., IoT time series) \cite{thelen2022comprehensive}.
In the first case, the concept ``entity'' is determined by the observational granularity level; e.g., it could be an electronic stability program (ESP) module or the whole powertrain of a car that contains the ESP. In the second case, the problem is (over-) simplified to finding the relationship between multivariate input from IoT sensors, which represents the physical status, and the target values under prediction; yet, the rich information contained in sensor data from communication networks is severely underutilized. 
% Even with sometimes extra consideration of both time and frequency domain, existing approaches focus mainly on the linear or non-linear relationship between sensor readings and unobservable target values \cite{thelen2022comprehensive}.

Moreover, current digital twin research cannot model the heterogeneous and complex relationship among the different physical entities in a unified manner. Various models have to be used to represent different relations and heavily rely on domain expertise \cite{thelen2022comprehensive}. Thus, the physical-to-virtual modeling for both entities and their relationships is often performed from scratch on an application-by-application basis. This is both costly and time-consuming, and renders it almost infeasible to construct a scalable digital twin with a large number of heterogeneous entities. These barriers motivated us to create a general data structure associated with a processing framework, that can model heterogeneous physical entities and their relationships without relying on domain knowledge (but rather on data from IoT sensor networks \cite{comlet12sdn}). In a nutshell, this paper makes the following contributions:

\begin{enumerate}
\item We propose Digital Twin Graph (DTG), a general graph representation of interconnected physical systems, which requires no domain knowledge and is completely data-driven.
%each physical entity, and model the relationship between every two possibly interacting entities using a Graph2Graph transformation model. 
Note that DTG is {\em not} a traditional graph neural network (GNN), where each arc is a simple weight which can only describe very simple relationship (linearly proportional) and cannot describe the complex physical world. On the other hand, our DTG uses a {\em Graph-to-Graph transformation model} that we propose.

\item We propose a generative adversarial ensemble network (GAEN) that fuses {\em existing} domain knowledge-based models with our data-driven models. This allows DTG to opportunistically {\em leverage} domain knowledge 
{\em as and when available}, thereby enhancing the fidelity of our digital twin.

\item We propose a GNN-based system-wide simulation strategy, allowing DTG to perform complex system simulation and updates about the real-time changes of any entity status or system structure.

\end{enumerate}

\begin{figure}[t]
\centering
\includegraphics[width=0.92\linewidth]{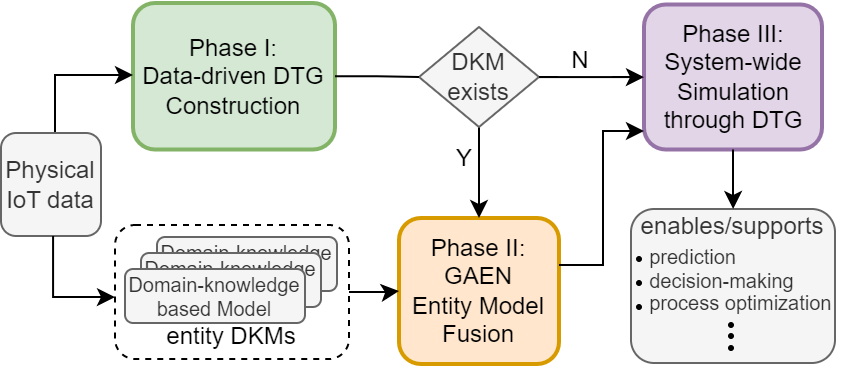}
\vspace{-2mm}
\caption{The processing framework of Digital Twin Graph.}
\label{DTG_architecture}\vspace{-5mm}
\end{figure}

\vspace{-2mm}

\section{DTG and Technical Challenges}

\begin{figure*}[t]
\centering
\includegraphics[width=0.94\textwidth]{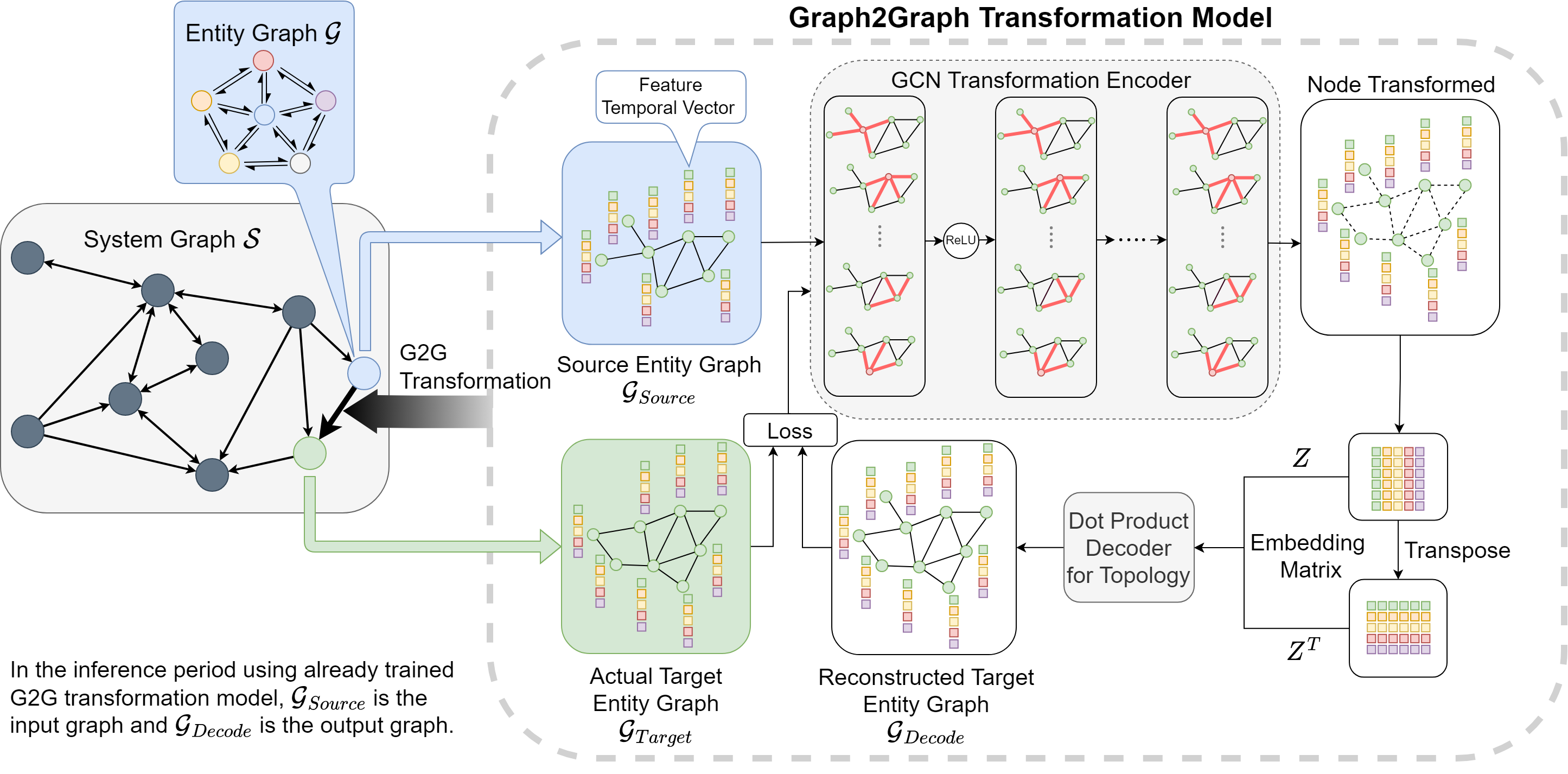}
\caption{The Digital Twin Graph structure and its associated G2G transformation model.}
\vspace{-3mm}
\label{DTG_construction}
\end{figure*}

The overall DTG framework is depicted in \fref{DTG_architecture}. First, IoT data from physical entities are fed to {\em Phase I: Data-driven DTG Construction}. Second, if there are {\em existing} domain-knowledge based DT models (DKM) that were built for some entities, we incorporate them into the data-driven DTG in {\em Phase II: GAEN Model Fusion}. Third, given our obtained digital representation of the physical world, which is DTG, we perform {\em Phase III: System-wide Simulation} to study how the physical system would function and react in response to various events and changes.

\newcounter{savedenum}
\newcommand*{\saveenum}{\setcounter{savedenum}{\theenumi}}
\newcommand*{\resume}{\setcounter{enumi}{\thesavedenum}}

\noindent{\bf{Phase I.}} \emph{Data-driven DTG Construction:} This phase is concerned with how to automatically construct a domain-agnostic Digital Twin Graph based on the physical system's sensor data. There are two main challenges:
\begin{enumerate}
    \item How to represent each physical entity that has multiple features and how to update its features when a subset of its feature values change?
    \item How to model the relationship among different entities in the system?
    \saveenum
\end{enumerate}

\noindent{\bf{Phase II.}} \emph{Entity Model Fusion:} This phase deals with designing a method to merge a data-driven entity model with the DKMs (if available) of the same entity. The challenge is:
\begin{enumerate}
    \resume
    \item A Phase I data-driven entity model is likely distinct from the DKMs, and the DKMs likely have different structures among themselves too. How to accomplish model fusion under such heterogeneity and incompatibility?
    \saveenum
\end{enumerate}

\noindent{\bf{Phase III.}} \emph{System-wide Simulation:} Given a DTG constructed through Phase I and II, the next question is how to utilize this DTG to simulate possible or intended events and changes as if in the real physical world. Specially,%which further can help system prediction, decision-making, and process optimization.
\begin{enumerate}
    \resume
    \item How to simulate the (direct or indirect) impact of various changes in any entities on other entities and the entire system through the graph structure?
    \item Given the various impact, how to keep the status of each entity and the whole system up to date with high fidelity?
\end{enumerate}

\section{Proposed Solution}

\noindent{\bf{Phase I: Data-driven DTG Construction}} 

A DTG is a graph representation of a physical system consisting of a (possibly large) number of physical entities that are related to each other in different ways. DTG models each physical entity as an {\em entity graph}, as elaborated in \sref{sec:entgr} which solves \textbf{\underline{Challenge 1}}, and models the whole physical system as a {\em graph of entity graphs}, as described in
\sref{sec:sysgr} which solves \textbf{\underline{Challenge 2}}. \fref{DTG_construction} gives an overview. The idea was inspired by \cite{beyond19icdm} where a (small) grid in a city map is represented by a POI graph and a larger geo-region is represented by a graph of grids each with concatenated POI features.

\vspace{-2mm}
\subsection{Entity Graph Construction}\label{sec:entgr}

An entity graph is constructed automatically for each entity based on the IoT sensor data about that entity, of which each feature corresponds to a sensor modality (temperature, illumination, humidity, pressure, vibration, torque, displacement, etc.). 

An entity graph is depicted as the blue block on the upper-left panel of \fref{DTG_construction}. We denote each entity graph by \(\mathcal{G}_{k}, 1\leq k \leq N_e\) where \(N_e\) is the total number of entities in the system. Each node in an entity graph \(\mathcal{G}_k\) represents a feature of that entity, denoted by \(V_{i}^k,  1\leq i \leq N_{f}^k\), where \(N_{f}^k\) means the number of features of the entity \(k\). Each arc in \(\mathcal{G}_k\) is denoted by \(E_{ij}^k,  1 \leq i,j \leq N_{f}^k\), where \(E_{ij}^k \in [-1,1]\) is the correlation coefficient between feature \(i\) and \(j\) in entity $k$; \(E_{ii}^k=1\). To choose a proper correlation method (e.g. Pearson's coefficient) for each particular entity, we conduct significance tests (e.g. t-test) and pick the correlation method that yields the highest confidence averaged over all feature pairs. After constructing an entity graph, we associate each of its features with a regression model over that feature's other correlated features which are determined by the structure of the entity graph $\mathcal{G}_{k}$. Each regression model will be used (in simulation) to update the value of its associated feature $V_i^k$ when any of its correlated features changes value. 

%Without changing the internal physical topological structure of all components in an entity, the virtual entity graph topology should remain the same.  This data structure can help solve the \textbf{\underline{challenge 1}} in lot of digital twin scenario in a general way. 

\vspace{-2mm}
\subsection{System-wide Graph Construction}\label{sec:sysgr}

Given all the entity graphs constructed from above, here we construct a system-wide graph which completes the DTG. The DTG is denoted as \(\mathcal{S}\) as depicted on the left panel of \fref{DTG_construction}, where each node is a entity graph \(\mathcal{G}\) constructed in \sref{sec:entgr}. 

The key to building an $\mathcal{S}$ is on the relationship between entity graphs $\mathcal{G}$. DTG uses a directed arc to represent each such relationship between two \(\mathcal{G}\)'s and model it using a {\em Graph-to-Graph (G2G) transformation model}, denoted by \(M_{ij},  1 \leq i,j \leq N_e, i \neq j,\) where $i$ and $j$ are two physically connected or directly related entities. Unlike in a conventional GNN, node relationship is modeled as a simple scalar (decay rate), DTG uses G2G transformation model which can characterize complex relationship between physical entities with high fidelity.

A G2G transformation model is depicted in the right panel of \fref{DTG_construction}, which models the mapping from a {\em source} entity graph $i$ (blue) to a {\em target} entity graph $j$ (green). More specifically, a G2G model serves the purpose of automatically updating the feature values of the target graph whenever the feature values of the source graph change. This updating rule generally varies between different source-target pairs, and hence we take a data-driven approach to {\em learn} this rule based on historical data.

%The two entity graphs need to have \mathcal{G} should be with fixed graph structure, meaning the entity inner structure stays stable. Therefore, as time goes, only the feature sensor values might change and cause the connected entity status changes, but not the structure inside the entities.

To train such a rule, we cast this transformation as a {\em reconstruction (encoding and decoding)} problem from $\mathcal{G}_{Source}$ (blue) to $\mathcal{G}_{Target}$ (green). We choose {\em Graph Convolutional Network} (GCN) as the encoder, because the convolutional architecture is similar to the first-order approximation of (spectral) graph convolutions, which can help transform each (temporal) feature vector (akin to the RGB channel in CNN) of a vertex in the source graph into one in the target graph. Through the GCN encoder and a Dot Product decoder, we will obtain a reconstructed graph $\mathcal{G}_{Decode}$ (see \fref{DTG_construction}) that is similar to the target entity graph $\mathcal{G}_{Target}$. We will then retrain $\mathcal{G}_{Decode}$ iteratively using a loss function to make $\mathcal{G}_{Decode}$ close enough to $\mathcal{G}_{Target}$.

For the encoding process (upper part of the G2G model as in \fref{DTG_construction}), we construct an adjacency matrix $\mathbf{A}$ from $\mathcal{G}_{Source}$, where each element $A_{ij} = \mathbbm{1}_{|E_{i,j}|>\delta}$, $\delta\in (0,1)$, so that the adjacency matrix captures feature pairs that are sufficiently correlated. %(e.g., $\delta=0.5$ is recommended). 
Likewise, we construct another adjacency matrix $\mathbf{A^\prime}$ from $\mathcal{G}_{Target}$ the same way. % (diagonal elements already set to 1, i.e.~every node is connected to itself)
Then, we introduce a feature matrix $\mathbf{X}$ for the source entity graph and the corresponding latent feature matrix $\mathbf{Z}$, where $\mathbf{X}$ is of dimension $N\times T$ and $\mathbf{Z}$ is of $N\times F$, $N$ is the number of features of the entity (i.e., simplifying the notation $N_f^k$), $T$ is the length of each temporal feature vector (i.e., the length of history we look back for each feature), and $F$ means the same as $T$ but {\em after} encoding. Thus, our GCN encoder is parameterized as follows:
\begin{align}
%\textstyle
\mathbf{H}^1 &= ReLU(\mathbf{A}\mathbf{X}\mathbf{W}^0)\\
 \mathbf{H}^i &= ReLU(\mathbf{A}\mathbf{H}^{i-1}\mathbf{W}^{i-1})
\end{align}
where $\mathbf{H}^i$ and  $\mathbf{W}^i$ are the output and the weight matrix, respectively, of the $i$-th layer, $i=0,1,2,...,l-2$ (layer-0 is input). Thus, a $l$-layer GCN is defined as:
\begin{equation}
\textstyle
  \mathrm{GCN}(\mathbf{X}, \mathbf{A}) = \mathbf{H}^{l-1}= \mathbf{A}\mathbf{H}^{l-2}\mathbf{W}^{l-2} = \mathbf{Z}.
  %\mathbf{A}ReLU\bigl(\mathbf{A}\mathbf{X}\mathbf{W}^0\bigr)\mathbf{W}^1
\end{equation}
%where $\mathrm{ReLU}(\cdot)=\max(0,\cdot)$ as the activation function.
% In  the decoding part, depending on the sparsity of entity graphs, we designed two pairs of decoder and loss function to better suit digital twin data scenarios. 
% \subsubsection{Relatively \textbf{SPARSE}} 
% If we are facing relative sparse entity graphs, 
During this encoding process, the topology described by $\mathbf{A}$ and the node features of the source entity are both encoded through first-order approximation into the latent feature matrix $\mathbf{Z}$. %while the topology corresponding to\red{"embedded by"?} $\mathbf{Z}$ is generally different from the topology of the target graph. 
We then extract the embedded topology according to the feature proximity described by $\mathbf{Z}$, to obtain $\mathcal{G}_{Decode}$. Formally, our topology decoder reconstructs the adjacency matrix $\mathbf{\hat{A}}$ of $\mathcal{G}_{Decode}$ by computing (see \fref{DTG_construction}):
\begin{equation}
\mathbf{\hat{A}} = \Bigg[ \hat{A}_{ij} \Bigg| \hat{A}_{ij} = \mathbbm{1}_{\sigma\left(\mathbf{Z} \mathbf{Z}^T \right)_{ij} > \delta} \Bigg]_{N\times N}, 
\label{eq:gae}
\end{equation}
where %$Z^T$ is transpose matrix of $Z$ and 
$\sigma(\cdot)$ is the logistic sigmoid function.

In order to make $\mathcal{G}_{Decode}$ close to $\mathcal{G}_{Target}$, we establish a loss function below to ensure similarity between them on both topology and node features.

\subsubsection{\textbf{Topology}} We use a {\em weighted binary cross-entropy loss} as the topology reconstruction loss between the decoded $\mathbf{\hat{A}}$ and the target $\mathbf{A^\prime}$. Our topology loss function is defined as:
\begin{equation}
\textstyle
    \mathcal{L_T} = 
    \sum \limits_{i,j=0}^{N-1} 
    % \sum \limits_{j=0}^{N-1}
    \left( - S\times A^\prime_{ij}\log(\hat{A}_{ij})
    -(1-A^\prime_{ij})\log(1-\hat{A}_{ij}) \right)
\end{equation}
where $S = \frac{N^2-|D|}{N^2}$ and $|D|=\sum_{i=0}^{N-1}\sum_{j=0}^{N-1} A^\prime_{ij}$ (the source and target graphs share the same $N$, which we explain in \sref{sec:adapt}).  Thus, $S\in[0,1]$ characterizes the {\em sparsity} of the adjacency matrix $A'$ since $N^2$ is the maximum possible number of arcs (recall that $E_{ii}=1$). Hence, adding $S$ before the first term stresses on {\em connected} arcs when $S$ is larger (a sparser graph), thereby mitigating extreme (too sparse or too dense) graph topologies. The reason to avoid extremes is that such matrices are prone to neglecting the under-represented class (connected or disconnected arcs) and hence can hardly learn the class members.

\subsubsection{\textbf{Node Features}} 
For this we define the following loss function:
\begin{equation}
\textstyle
    \mathcal{L_N} = 
    \sum \limits_{f=0}^{N-1} \|\mathbf{x}_f - \hat{\mathbf{x}}_f\|_2
    % =\sum_{i,j}^{N} (\hat{A}_{ij}-A^\prime_{ij})^2
\end{equation}
where $\mathbf{x}_f$ is the feature vector of node $f$ in $\mathcal{G}_{Target}$; likewise is $\hat{\mathbf{x}}_f$ for $\mathcal{G}_{Decode}$.

Therefore, our Target-Decode loss for both graph topology and node features is defined by
\begin{equation}
    \mathcal{L_{TD}}= \mathcal{L_T}+\lambda \mathcal{L_N}
\end{equation}
where $\lambda$ is a hyperparameter allowing us to adjust priority between the two losses. By training the entire G2G transformation model using $\mathcal{L_{TD}}$ over source and target entity graphs generated from Phase I, we can learn parameters $\mathbf{W}^i$ of the GCN encoder.

% \subsubsection{Relatively \textbf{DENSE}}

% If we are facing relative dense entity graphs, we get the reconstructed adjacency matrix $\mathbf{\hat{A}}$ by:
% \begin{equation}
% \mathbf{\hat{A}} = \mathbf{Z} \mathbf{Z}^\top \, 

% \label{eq:f-loss}
% \end{equation}where $Z^T$ is the transpose matrix of $Z$ and there is no hyperbolic tangent function.

% For non-binary reconstructed entity graph, given two graphs $\mathcal{G}_{Decode}$,$\mathcal{G}_{Target}$ of the same order $N$ with correlation matrices $\hat{A}$, $A^\prime$, a well-studied measure of similarity is the Frobenius distance \cite{grohe2018graph}:
% \begin{equation}
% \label{deqn1}
% % x = \sum_{i=0}^{n} 2{i} Q.
% dist(\hat{A}, A^\prime)=\|\hat{A} - A^\prime\|_F=\sum_{i,j}^{N} (\hat{A}_{ij}-A^\prime_{ij})^2
% \end{equation}

% Where i,j range over all permutations from 0 to N-1 of the  vertex set of $\mathcal{G}_{Decode}$, where $A_{Dij}$ denotes the matrix obtained from $A_{D}$ by permuting rows and columns according to i,j, and $A_{Tij}$ is similar, where$\|M\|_F$ is the Frobenius norm of a matrix M. We define the loss  function F-Loss $L_{f}$ of Graph2Graph Transformation Model as 
% \begin{equation}
% \label{deqn2}
% % x = \sum_{i=0}^{n} 2{i} Q.
%  \mathcal{L_F} =\frac{dist(\hat{A},A^\prime)}{N^2}
%                =\frac{\sum \limits_{i=0}^{N-1} 
%                       \sum \limits_{j=0}^{N-1} 
%                       (\hat{A}_{ij}-A^\prime_{ij})^2}{N^2}
% \end{equation}

% since the combination number of $i,j$ is $N^2$ and we keep training until the loss is minimized to a threshold or curtain time limit. 
\vspace{-2mm}
\subsection{Graph Adaption}\label{sec:adapt}

The G2G transformation model requires each source-target graph pair to have the same order (i.e., the same number of features). This is because GCN can add and delete arcs but not nodes in a graph, during the encoding process. Therefore, when the two entities have different numbers of features, we need to adapt the entity graphs as follows.

\begin{itemize}
    \item $N_{f}^{Source} < N_{f}^{Target}$:
In this case, we pad pseudo features and random arc weights to the source entity graph. 
Each pseudo feature has a $\mathbf{0}$ feature vector and has an {\em inward} connection from each of the other {\em real} features, so that these padded features will use real feature vectors to compose their feature vectors during the encoding process (into $Z$), while not affecting either the real features or the target entity features. Thus ultimately, the $\mathbf{0}$ feature vectors will not contribute to the transformation of node features or the corresponding topology, and the finally trained $\mathcal{G}_{Decode}$ and the parameters $\mathbf{W}^i$ of the GCN encoder will not be affected by the pseudo features in the source entity graph.

\item $N_{f}^{Source} > N_{f}^{Target}$:
In this case, we run K-means clustering on the source entity graph with $K=N_{f}^{Target}$ so that strongly correlated features in $\mathcal{G}_{Source}$ merge into ``hyper'' features, represented by cluster centroids. % \blue{Because the Target gragh will be part of the Loss function, and we can't introduce this kind of manual bias without good reasons}
\end{itemize}
One may think we could, reversely, perform clustering on $\mathcal{G}_{Target}$ in the first case or add pseudo features to $\mathcal{G}_{Target}$ in the second case as a plausible alternative. However, this will not work because $\mathcal{G}_{Target}$ is the calibration benchmark for the G2G transformation and should not be changed; otherwise some deviation will be introduced into the loss function. %bias, aberration, diversion, distortion
% Besides, if some entity graph is very sparse, meaning the  features are very independent to each other, we will train the DTG arc model related to that arc as a vector-to-vector model, with the snapshot of all feature values at one time point as a training data point. The feature values are concatenated as vectors. And different vector pairs among two entities along the temporal perspective will be the whole training dataset. 
At this point, \textbf{\underline{Challenge 2}} is solved, in a domain-agnostic manner.

\vspace{\baselineskip}
\noindent{\bf{Phase II: Entity Model Fusion}} 

\begin{figure}[t]
\centering
\includegraphics[width=0.96\linewidth]{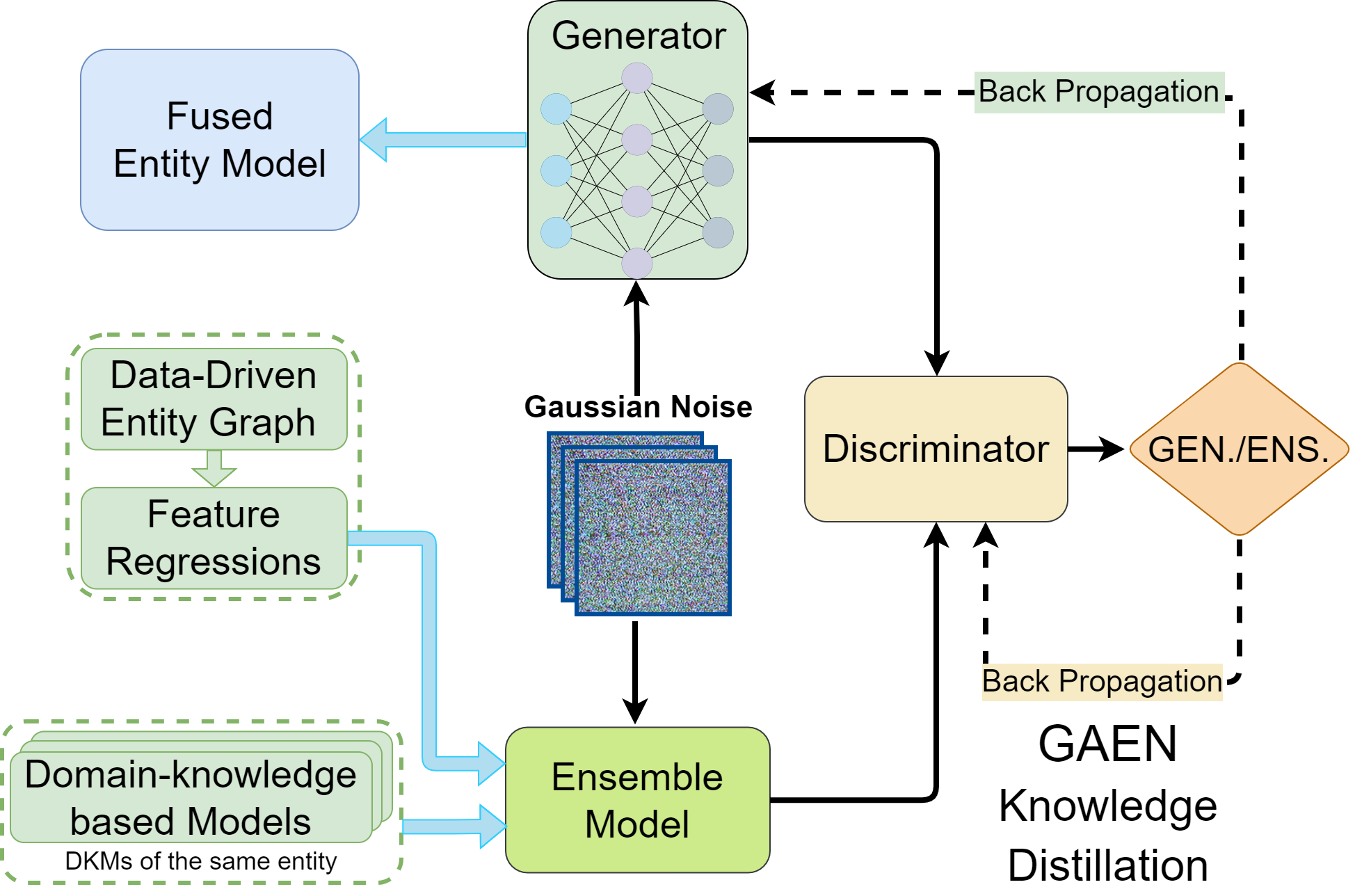}
\vspace{-2mm}
\caption{Entity Model Fusion using GAEN.}
\vspace{-2mm}
\label{GAEN_Fusion}
\end{figure}

In practice, some entities (e.g. key components in vehicles, robots in manufacturing plants) have already been digitized in high fidelity domain-knowledge based models (DKMs) with their complete or partial set of features. Hence, it will be helpful to leverage these DKMs rather than relying on a data-driven approach exclusively. To this end, we propose a model fusion method called Generative Adversarial Ensemble Network (GAEN) based on {\em knowledge distillation}. 

As depicted in \fref{GAEN_Fusion}, our GAEN takes an entity graph (more specifically all of its associated feature regression models; cf. \sref{sec:entgr}) and the (potentially multiple, heterogeneous) DKM(s) of the same entity as input. GAEN first trains an ensemble model to represent the mixture of all the input models. As a result, the ensemble model may end up with a large number of parameters and hence increases its risk of overfitting. Furthermore, due to the potentially numerous entities in a physical system, the entire DTG may contain a massive number of parameters due to the ensembles models. These may jeopardize the scalability of a comprehensive digital twin in its construction and simulation processes. 

To address this issue, we create an equivalent generative adversarial network, that has much fewer parameters, for each ensemble model. We utilize a generator to take in random Gaussian noise and make prediction of all the feature values of that entity. Meanwhile, the same random noise will also be fed to the ensemble model to make that prediction too. Then, we use a discriminator to differentiate between the result of the generator and that of the ensemble model, and backpropagate the difference to the generator. This process iterates until the discriminator fails to tell the difference between the outputs of the two models. Upon this point, the generator is a good surrogate and lightweight model of the ensemble model where it (the generator, i.e., the student model) has distilled knowledge from both our data-driven entity graphs and the available DKMs (i.e., the teacher models). This tackles \textbf{\underline{Challenge 3}}.

\vspace{\baselineskip}
\noindent{\bf{Phase III: System-wide Simulation}} 
% \blue{Change to regular sim + G2G changing situation(update procedure+example case type), downplay the "structure changing or not" concept}

%In the phase of system-wide simulation, there are two scenarios to handle depending on whether the physical system structure (i.e., topology and entities' ...) is changed.

The most typical scenario for system-wide simulation is to handle status updates of the interconnected entities. Generally, only some entities' status is observable (i.e., their IoT sensor data are fed into the digital twin in time) while others are not (at that moment) but need to be updated as well (as a result of other entities' status change) so as to reflect the most up-to-date status of the entire system.

% \red{delete this para}In normal status, when the whole physical system is running as designed, IoT sensors from each entity and updates the entity feature values. But not all entities are observable in real time, some of them can only be observed periodically due to operation restriction or cost. Between those observations, we still need to update  all entity status in the Digital Twin according to the observable ones. 

% \red{this para basically talks about the same scenario; delete it}In environment abnormal status, when certain entities have encountered feature values changes due to those entity environment changes, we also need to update the newest entity status and run the simulation across the whole system, so that we can observe the impact of those entity status changes.\blue{It is meant to another situation: The DTG graph structure doesn't change, but the environment for certain entities changed, so the whole system will respond to the environment change, which needs simulation for the system-wise reaction.}\red{this is still the 1st scenario, where entity status changed; of course such change may be caused by environment change or itself change, but the outcome is the same, so still can delete.}\blue{That's true, just the cause difference}

In the other scenario, the physical system undergoes structural changes or the virtual twin is simulating structural changes, e.g., entity breakdown or replaced by a different type, new entities added or connections between entities broken/replaced/added. In such cases, the originally trained G2G transformation models for those affected entities can no longer be used for inference.

To handle the first scenario, we utilize our proposed G2G transformation models to propagate the status change of observable entities to the whole system graph until reaching a new stable status (i.e., converged). Note that our G2G models are radically different from existing GNNs whose propagation rule is a simple decay rate (a scalar), whereas our rule as expressed by G2G can capture much more complex relationship between real entities.
% \red{delete this sentence}When these two problems happen, both physical entities internal structure and the system structure connecting all entities are unchanged, which means the entity graphs and G2G models between entity nodes remain the same. 
% \red{the following comes from nowhere; why we suddenly need a VGAE? the scenario seems to have been handled already.}\blue{VGAE was used in the first version of simulation, which is a candidate of GNN-base strategy, I think we can delete this part now}Then we can utilize Variational Graph Auto-Encoders \cite{kipf2016variational} to learn the system graph embedding, and further enable or benefits prediction and decision-making depicted in \fref{DTG_architecture}. 

To handle the second scenario, we need new G2G models and new entity graphs to replace those affected ones. This can be achieved by repeating our Phase I but only for the affected area instead of reconstructing the whole system graph from scratch. In fact, this process can be expedited even further, by creating a DTG database that keeps a copy of all entity graphs and G2G models once they are constructed or (pre-) trained (which can be done offline). Thus when a new entity graph or G2G is needed, we first query the DTG database to retrieve the same object if it already exists, and only train a new one otherwise. We call this a {\em lazy-training} strategy.
% \blue{feels similar to \em lazy-learning, and I like this saying}
After replacing the G2G models, this second scenario become the first scenario and we simply run the same procedure thereof to update the entire DTG.
Upon this point, we have addressed \textbf{\underline{Challenge 4}} and \textbf{\underline{Challenge 5}} through the graph propagation process.

\vspace{-2mm}
\section{Discussion and Conclusion}
% In Phase I, the correlation models in different entities varies a lot, which impact the feature prediction greatly. Thus, if certain side information is available, we should classify the entities first and adapt correlation models according to the classification. Regarding the arc modeling, it is also possible in a pair of connected entities, their entity graph has very little connection. It means only a few features are affected by the adjacent entity status changes.  If this isolation happens, approach to detect it should be studied in order to prevent arc model training continuously without result, either in the train session delicately design the loss function or before training explore the graph-to-graph models.

DTG allows extra flexibility where, in Phase II, we can assign weights to DKMs when building the ensemble model to reflect our confidence or trust in the DKMs. In Phase III, one may be concerned about computational complexity, but in reality, a DTG is unlikely to have many strongly connected components (SCC); in fact, it is often rather sparse. Moreover, a digital twin is typically run in an HPC cluster that has ample computing resources.
%In fact, we can even keep some of the entity graphs in edge servers of wireless sensor networks, which can further alleviate the computation load via a {\em distributed digital twin}.

In this paper, we propose DTG as a general data structure for automated Digital Twin construction and simulation, empowered by G2G transformation and GAEN model fusion. With an end-to-end three-phase processing framework, we made an effort to push the research in Digital Twin to a new stage where excessive domain knowledge is no longer required, which would enable and accelerate large-scale Digital Twin development based on IoT data.

% Besides, Ensemble model is not the only option for the model fusion. And for the task of unsupervised learning, we still need to explore other solutions.
% if the connection pattern between entities are totally changed, we can retrain only neural network predictor related to those arcs which is very time saving and potentially capable of being finished in real-time manner.

% One major short back of DTG and this framework is it can not handle new physical system very well since the nature of all models are data-driven, which means it is hard to deploy the digital twin the same time with brand new physical system. This issue might need the help of research in problem of  transfer learning with cold start.

% The current work we are implementing is focusing on static status of the system, which utilize majorly history system status as separate snapshots. By nature, it is spatial or structural causality. While there are plenty to rich information in the temporal dependency perspective. Thus, we will expend the framework to next level in the following research,  which fuse the temporal pattern or temporal causality into this framework, making it more general for more scenarios and disciplines.

\vspace{-2mm}
\bibliographystyle{ieeetr}
\bibliography{bibtex}
% \begin{thebibliography}{1}

% \bibitem{gelernter1993mirror}Gelernter, D. Mirror worlds: Or the day software puts the universe in a shoebox... How it will happen and what it will mean. (Oxford University Press,1993)
% \bibitem{grieves2019virtually}Grieves, M. Virtually intelligent product systems: digital and physical twins.  (2019)
% \bibitem{piascik2012materials}Piascik, B., Vickers, J., Lowry, D., Scotti, S., Stewart, J. \& Calomino, A. Materials, structures, mechanical systems, and manufacturing roadmap. {\em NASA TA}. pp. 12-2 (2012)
% \bibitem{DOD1998digital}Office of the Under Secretary of Defense, Research and Engineering (USD (R and E)): ``Digital Engineering''. https://ac.cto.mil/digital-engineering/, (accessed Jan. 01, 2023)
% \bibitem{thelen2022comprehensive}Thelen, A., Zhang, X., Fink, O., Lu, Y., Ghosh, S., Youn, B., Todd, M., Mahadevan, S., Hu, C. \& Hu, Z. A comprehensive review of digital twin—part 1: modeling and twinning enabling technologies. {\em Structural And Multidisciplinary Optimization}. \textbf{65}, 1-55 (2022)
% \bibitem{grohe2018graph}Grohe, M., Rattan, G. \& Woeginger, G. Graph similarity and approximate isomorphism. {\em ArXiv Preprint ArXiv:1802.08509}. (2018)
% \bibitem{kipf2016variational}Kipf, T. \& Welling, M. Variational graph auto-encoders. {\em ArXiv Preprint ArXiv:1611.07308}. (2016)
% \end{thebibliography}

\end{document}